\documentclass[letterpaper, 10 pt, conference]{ieeeconf}  

\IEEEoverridecommandlockouts                              

\usepackage{cite}
\usepackage{amssymb}
\usepackage{amsmath}
\usepackage[table]{xcolor}
\usepackage{multirow}
\usepackage{graphicx}
\usepackage{xspace}
\usepackage{subcaption}
\usepackage{caption}
\usepackage{floatrow}
\usepackage{multicol}
\usepackage{dsfont}
\usepackage{booktabs}
\usepackage{multirow}

\usepackage[normalem]{ulem}

\title{Self-supervised Wide Baseline Visual Servoing via 3D Equivariance}
\author{Jinwook Huh, Jungseok Hong, Suveer Garg, Hyun Soo Park, and Volkan Isler 
\thanks{All authors are with the Samsung AI Center NY, 837 Washington St, New York, NY 10014}%
}
\date{February 2022}

\begin{document}

\maketitle


\begin{abstract}
One of the challenging input settings for visual servoing is when the initial and goal camera views are far apart. Such settings are difficult because the wide baseline can cause drastic changes in object appearance and cause occlusions. 
This paper presents a novel self-supervised visual servoing method for wide baseline images which does not require 3D ground truth supervision. Existing approaches that regress absolute camera pose with respect to an object require 3D ground truth data of the object in the forms of 3D bounding boxes or meshes. We learn a coherent visual representation by leveraging a geometric property called 3D equivariance---the representation is transformed in a predictable way as a function of 3D transformation. 
To ensure that the feature-space is faithful to the underlying geodesic space, a geodesic preserving constraint is applied in conjunction with the equivariance. 
We design a Siamese network that can effectively enforce these two geometric properties without requiring 3D supervision. With the learned model, the relative transformation can be inferred simply by following the gradient in the learned space and used as feedback for closed-loop visual servoing. Our method is evaluated on objects from the YCB dataset, showing meaningful outperformance on a visual servoing task, or object alignment task with respect to state-of-the-art approaches that use 3D supervision. Ours yields more than $35\%$ average distance error reduction and more than $90\%$ success rate with $3$cm error tolerance.

\end{abstract}

\section{Introduction}

Camera motion introduces scene parallax, which provides a strong visual cue to, in turn, recover the camera egomotion. The ability to recover such camera egomotion brings out a number of robotics applications such as image-based auto-steering~\cite{du2019self,jayaraman2015learning}, hand-eye coordination for robot object tracking~\cite{costanzo2021can,devgon2020orienting}, and active scene perception for navigation. Classical visual servoing approaches~\cite{hutchinson1996tutorial,corke1996visual,deguchi2000direct} that leverage local image feature matching have shown remarkable performance, which enabled robots to precisely estimate the relative transformation without prior knowledge. A key challenge arises in wide baseline egomotion where local image feature matching does not apply which can be due to occlusions or the fact that the appearance of local image patches changes drastically as the baseline increases. 
Learning-based approaches which regress the absolute pose of the camera with respect to a pre-defined object coordinate system offer a viable remedy. With these methods, the relative transformation can be derived from two predicted absolute poses. These approaches, however, require 3D supervision: the 3D spatial relationship between the camera and pre-defined object coordinate system. The pre-defined coordinate system needs to be labeled in some form, such as a 3D bounding box or mesh~\cite{xiang2017posecnn}, which is difficult to scale for general objects. 

\begin{figure}[t]
    \centering
    \includegraphics[width=\columnwidth]{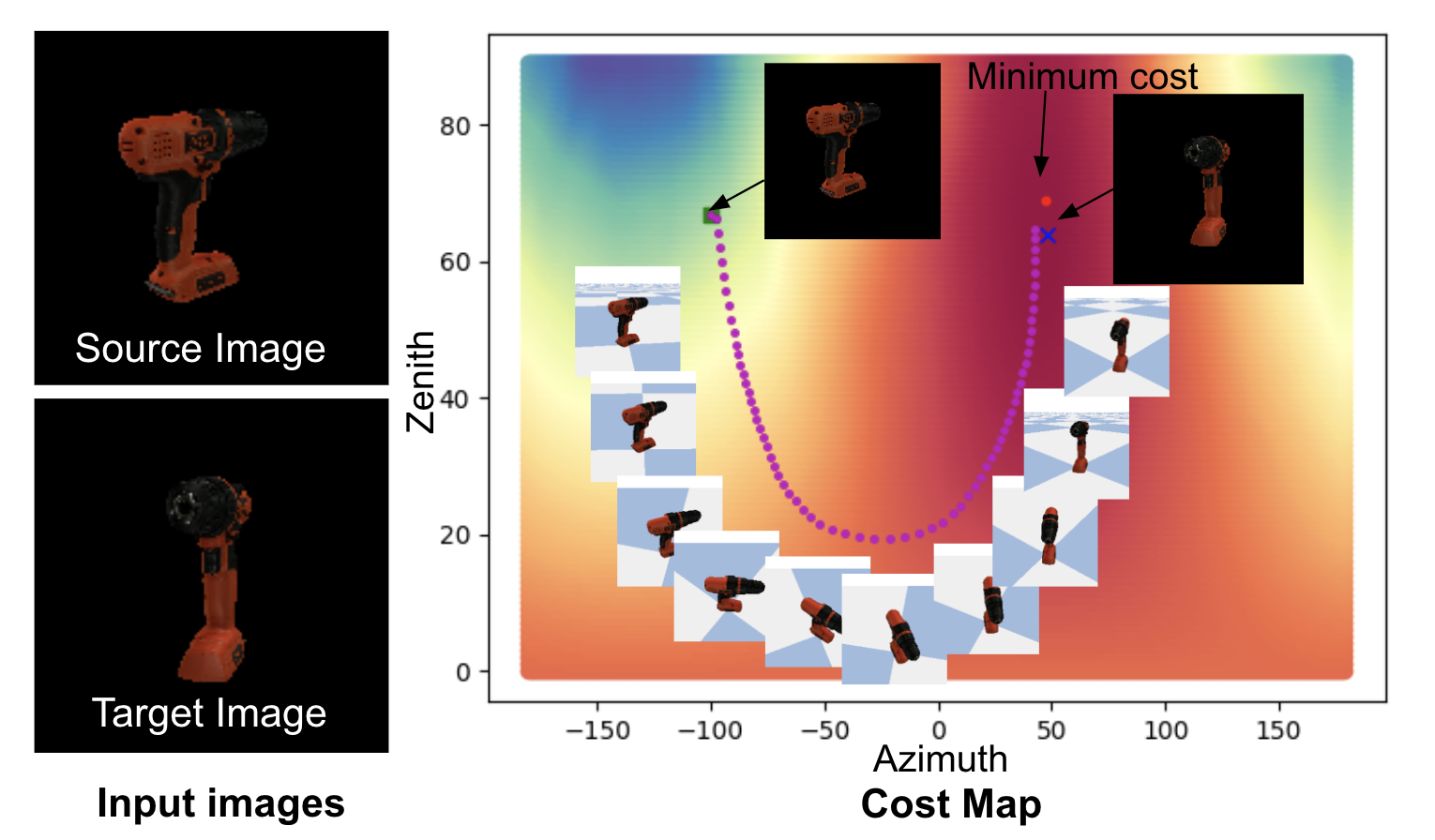}
    \caption{We present a self-supervised learning method for a wide baseline visual servoing task. Given the source (initial) image, we subsequently predict the relative transformations to the target image of which visual appearance is significantly different from the source image. We visualize the camera trajectory on the cost map in $\mathds{S}^2$. The value in the cost map indicates the distance in the learned visual representation. }
\label{fig:system_overview}
\end{figure}

In the context of visual servoing, there are significant recent efforts to address the limitations of classical methods using learning-based approaches~\cite{bateux2017visual,yu2019siamese,puang2020kovis,devgon2020orienting}. 
At the moment, these methods rely on direct regression of the absolute pose as input to the controller or learning of visual keypoints. In this paper, we show that visual servoing can be performed directly in the learned feature space. Specifically, we present a novel self-supervised visual servoing method for wide baseline transformations without 3D supervision. Our main idea is inspired by human embodied perception, where we learn object transformation as we move without 3D ground truth. Specifically, there exists a visual representation that possesses a geometric property called \textit{3D equivariance}---the visual representation is transformed as a function of 3D transformation. For instance, when an object is translated from one location to another, its visual representation must undergo the transformation that accounts for the translation. (Fig.~\ref{fig:system_overview})

Our method uses pairs of source and target images and their relative transformation to learn a coherent visual representation. To avoid a trivial solution, we integrate the geodesic preserving property into representation learning in conjunction with 3D equivariance. With the learned representation, we formulate the problem of estimating the optimal transformation as one can best explain the feature transformation. 


To learn a generalizable representation, our approach needs the training dataset that spans all possible combinations of source and target pairs over $SE(3)$ manifold. This is, in practice, highly intractable due to its six-dimensional nature. We introduce dimensional reduction by object centering scheme, i.e., an object is centered at the image center (principle point) with fixed size as if the camera closely tracks the object with constant distance. This reduces the space of transformation to three dimension, which allows representation learning with tractable sampling. 

We design a Siamese network that can effectively apply 3D equivariance and geodesic preservation. Each network is made of the visual feature extractor (ResNet18), and the feature from the source image is transformed by the feature transformer (multilayer perceptrons) conditioned on the input transformation. We evaluate our method on YCB data by comparing with state-of-the-art learning-based visual servoing approaches. Although our method is not designed for 3D alignment, we show that it also outperforms the others in 3D alignment metrics. 

\noindent \textbf{3D object supervision vs. 3D egomotion supervision} Our method uses 3D relative egomotion as a supervisionary signal to learn a representation. This supervision differs from 3D object supervision. In visual servoing, 3D camera egomotion can be trivially obtained by motor encoder readings unlike 3D object geometry. To attain 3D ground truth for object geometry, in practice, it requires (1) a 3D representation from either 3D CAD models or 3D reconstruction\footnote{An additional mesh or bounding box alignment is needed for multiple instances.} and (2) a visual alignment with images. This alignment is often conducted by manual annotation or sophisticated capture systems. Our approach uses the camera egomotion from encoder reading, which is readily generalizable to a new set of objects.

In summary, the core contributions of the paper are  1)~self-supervised learning for visual feature representation and equivariance feature transformation; 2) ~estimation of relative transformation for the target image with respect to a current camera coordinate frame using backpropagation of network; 3)~an improved servoing method which can handle large changes in target object appearance.

\section{Related Work}
Visual servoing (VS) is one of the core robot capabilities, applicable to a wide range of robotics tasks including robot navigation, tracking, and robot control with hand-eye cameras~\cite{hutchinson1996tutorial,corke1996visual}. Classical VS uses a measure of local image feature matching~\cite{{hutchinson1996tutorial,corke1996visual,deguchi2000direct}} as visual sensory feedback where visual alignment with sub-pixel precision can be achieved when the source and target images are similar enough, i.e., narrow baseline. However, the classical VS is not applicable to wide baseline visual servoing because the main assumption of visual feature matching does not hold, i.e., the local feature is valid only up to affine transformation. This leads to a local minimum for 3D transformation search. 

Camera pose estimation with respect to an object coordinate system has been studied, which can address the limitation of classical VS, e..g, using 3D reconstructed geometry of an object~\cite{mur2015orb,sattler2011fast,mur2017orb,sattler2016efficient} to localize camera poses. More flexible representation can be achieved by incorporating representation learning: regressing absolute camera pose directly from RGB or depth images~ \cite{kendall2015posenet,melekhov2017relative,brahmbhatt2018geometry}. Notably, Sattler et al. \cite{Sattler_2019_CVPR} extensively discussed the limitation of these end-to-end learning approaches.


On the other hand, in robotics, there are state-of-the-art object pose estimation approaches based on deep neural networks \cite{xiang2017posecnn,tremblay2018deep,hinterstoisser2012model,li2018deepim} for grasping with a robot manipulator. These approaches focus on pose regression in cluttered environments with 3D supervision (requiring 3D ground truth object bounding boxes or meshes). This requirement poses a substantial challenge because obtaining such ground truth data, in particular 3D data is difficult. Existing datasets provide a limited set of objects where application to generalized objects remains in question. Our method addresses this challenge by learning the relative transformation directly without 3D supervision. 



Devgon et al. \cite{devgon2020orienting} suggest a deep learning regression approach to estimate a relative orientation between two depth images. Since this approach can estimate the orientation within 30 degrees, it is hard to apply for a wide range of applications. Bateux et al. \cite{bateux2017visual} suggest a CNN-based visual servoing system. This approach iteratively regresses a relative pose from a current image to a given reference image. The network is trained in a supervised manner for regression of the relative pose between two input images. Samuel et al. \cite{felton2022visual} also propose an autoencoder network to learn low dimensional embedding space and perform VS in the latent space. However, they use generated images from only one scene or few scenes on the plane surface.
For the motion of cameras or robots based on visual inputs, there are many Reinforcement Learning (RL) approaches \cite{sampedro2018image, lee2017learning}. These RL approaches learn an inverse model from collected image-action sequences, and then the trained inverse model predicts action for given consecutive images \cite{habitat2020sim2real}. However, the performance of these approaches is degraded drastically if the input image is from out-of-distribution. In addition, RL approaches generate an action based on several consecutive visual inputs, which is hard to generate a whole sequence of trajectory to the target image.


\section{Method}
\label{section:method}
We estimate the relative 3D transformation between the source and target images where the baseline between two images (camera poses) is substantially large, i.e., local feature matching (SIFT~\cite{maxim2012robotic}) based approaches do not apply. Unlike existing works~\cite{xiang2017posecnn,tremblay2018deep,hinterstoisser2012model,li2018deepim} that learn absolute pose with respect to an object coordinate system, hence requiring labeled data, we design a self-supervised representation learning framework based on their relative transformation. To enable self-supervised learning, we leverage a geometric property called 3D equivariance to jointly learn a visual representation and its feature transformation in an end-to-end manner. To avoid a trivial solution, we incorporate a geodesic preserving constraint. Note that our method does not require any supervision for the object-specific coordinate system. To that end, we design a Siamese neural network architecture that can effectively enforce the 3D equivariance.

\subsection{3D Equivariant Representation Learning}
Consider an image $\mathbf{I}\in \mathds{I}$ where $\mathds{I} = [0,1]^{W\times H \times 3}$, and $W$ and $H$ are the width and the height of the image, respectively. The image is transformed to a target image via a warping function $\mathcal{W}:\mathds{I}\times \mathds{R}^d \rightarrow \mathds{I}$ parametrized by a transformation, i.e., $\mathbf{I}_{\rm tar} = \mathcal{W}(\mathbf{I}, \mathbf{p})$ where $\mathbf{I}_{\rm tar}$ is the target image, and $\mathbf{p}$ is a parametric transformation. For example, in a perspective image warping, $\mathbf{p}\in \mathds{R}^8$ is the homography, and $\mathcal{W}$ is the image compositional mapping. 

Equivariance is a geometric property that the visual feature of $\mathbf{I}$ is transformed in a predictable way:
\begin{align}
    f(\mathcal{W}(\mathbf{I},\mathbf{p})) = h_{\mathcal{W}}(f(\mathbf{I}))
\end{align}
where $f:\mathds{I}\rightarrow \mathds{R}^n$ is the learnable \textit{feature extractor} that takes as input the image and outputs $n$ dimensional feature vector. $h_{\mathcal{W}}:\mathds{R}^n\rightarrow\mathds{R}^n$ is the \textit{feature transformer}, parametrized by the warping $\mathcal{W}$. Feature invariance is the special instance of equivariance where $h_{\mathcal{W}}$ is the identity mapping, i.e., $f(\mathcal{W}(\mathbf{I},\mathbf{p})) = f(\mathbf{I})$.

We cast the visual servoing problem as a feature matching problem by generalizing 2D equivariance to 3D equivariance where $\mathbf{p}\in SE(3)$.  Specifically, we jointly learn the feature extractor $f$ and transformer $h_{\mathcal{W}}$ by minimizing : 
\begin{align}
    \mathcal{L}_{\rm equi}(\boldsymbol{\theta}_f,\boldsymbol{\theta}_h) &= \sum_{\{\mathbf{I}^i_{\rm src},\mathbf{I}^i_{\rm tar}\}_i} \|f(\mathcal{W}(\mathbf{I}^i_{\rm src},\mathbf{p}_i)) - h_{\mathcal{W}}(f(\mathbf{I}^i_{\rm src}))\|^2\nonumber\\
    &= \sum_{\{\mathbf{I}^i_{\rm src},\mathbf{I}^i_{\rm tar}\}_i} \|f(\mathbf{I}_{\rm tar}^i) - h_{\mathcal{W}}(f(\mathbf{I}_{\rm src}^i))\|^2, \label{Eq:equivariance}
\end{align}
where $\mathbf{I}_{\rm src}$ and $\mathbf{I}_{\rm tar}$ are the source and target images that are related by their relative $SE(3)$ transformation $\mathbf{p}$. $\boldsymbol{\theta}_f$ and $\boldsymbol{\theta}_h$ are the learnable weights that parametrize the feature extractor and transformer, respectively. 

\subsection{Geodesic Preserving Representation Learning}
Representation learning based on the equivariance alone may lead to a trivial solution, e.g., $f(\mathbf{I}) = \boldsymbol{0}$. We prevent such trivial cases by enforcing geodesic preservation:
\begin{align}
    d(h_{\mathcal{W}}(\mathbf{I}), f(\mathbf{I})) = c\|\mathbf{p}\|,
\end{align}
where $d(\cdot,\cdot)$ is the geodesic distance between two features, and $c$ is a non-zero constant. Since the feature distance between a feature and its transformed feature is proportional to the relative transformation between two, the learned representation is expected to span the space of $SE(3)$, i.e., nontrivial. 

We use Euclidean distance in $\mathds{R}^n$ as a distance metric, i.e., $d(\cdot,\cdot) = \|\cdot, \cdot\|$. With the distance metric, we enforce geodesic preservation by minimizing the following loss:
\begin{align}
    \mathcal{L}_{\rm geo}(\boldsymbol{\theta}_f, \boldsymbol{\theta}_h) = \sum_{\{\mathbf{I}^i\}_i} |d(h_{\mathcal{W}}(\mathbf{I}^i), f(\mathbf{I}^i)) - c\|\mathbf{p}\||.
\end{align}

Our self-supervised learning loss combines the equivariance and geodesic preservation:
\begin{align}
    \mathcal{L} = \mathcal{L}_{\rm equi} + \lambda \mathcal{L}_{\rm geo}
\end{align}
where $\mathcal{L}$ is the total self-supervised loss, and $\lambda$ is the weight that balances between the equivariance and geodesic preservation. Note that there is no notion of the object coordinate system which requires the labeled data (absolute pose with respect to an object). All formulations are based on relative transformation, which facilitates unsupervised learning. 

\begin{figure}[t]
\centering
\includegraphics[width=\columnwidth]{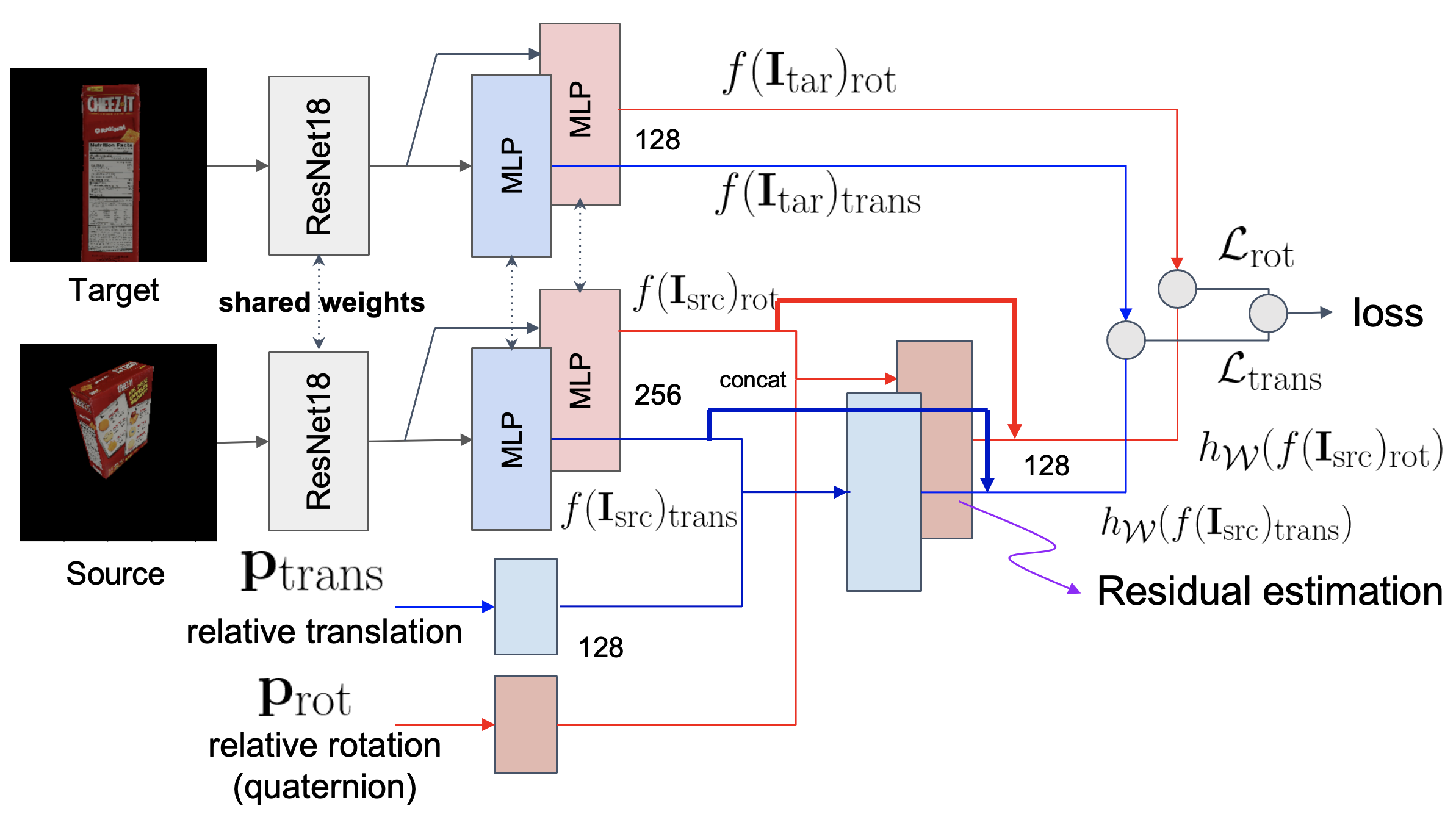}
\caption{We implement 3D equivariance using a Siamese network that is made of feature extractor $f$ and feature transformer $h_{\mathcal{W}}$. The equivariance is enforced by minimizing the feature difference between the target feature and the transformed source feature. MLP is used to model the feature transformer, parametrized by 3D transformation $\mathbf{p}$. }
\label{fig:total_network}
\end{figure}

\subsection{Dimensional Reduction by Object Centering} \label{Sec:obj_center}
\begin{figure}[t]
\centering
\includegraphics[width=\columnwidth]{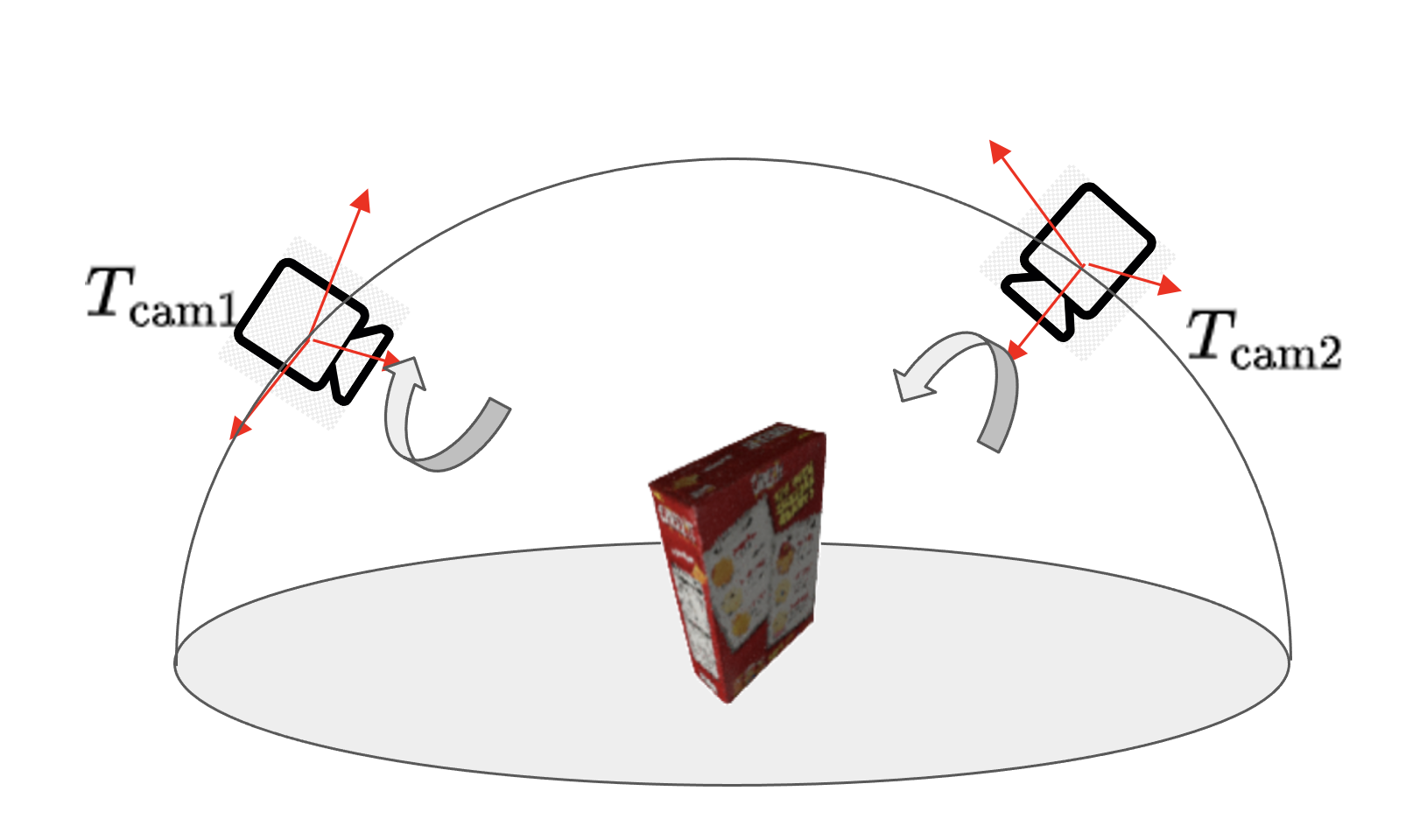}
\caption{Dataset collection. We generate images by uniformly sampling camera positions on the hemisphere space and the roll angle in the interval $[-\pi, \pi]$.}
\label{fig:data_capture}
\end{figure}

The parameterization of the transformation in Equation~(\ref{Eq:equivariance}) requires extensive sampling over all possible combinations of the pairs $\{\mathbf{I}_{\rm src},\mathbf{I}_{\rm tar}\}$, distributed over 3D rotation and translation (6 degree of freedoms of $\mathbf{p}$). Generating such a large dataset is challenging, which fundamentally limits applicability of equivariance based self-supervised learning. This makes a sharp contrast with object coordinate regression~\cite{xiang2017posecnn,tremblay2018deep,hinterstoisser2012model} that does not require pair data. To address this limitation, we reduce the dimension of transformation by centering an object on an image:
\begin{align}
    \overline{\mathbf{I}} = \mathcal{W}(\mathbf{I}, \mathbf{h}),
\end{align}
where $\mathbf{h}\in \mathds{R}^3$ is a homography that transforms an image at arbitrary pose $\mathbf{I}$ to the image of which principle coordinate is aligned with the object center with fixed object size as shown in Fig. \ref{fig:data_capture}:
\begin{align}
    \mathbf{h} = \mathbf{K}^{-1}\mathbf{S}\mathbf{R}\mathbf{K}\label{Eq:homography}
\end{align}
where $\mathbf{K}$ is the known intrinsic parameter of the camera, and $\mathbf{S} = diag(s,s,1)$ is the diagonal scale matrix where $s$ is the scale factor that changes the object size to a fixed size. The scale factor can approximate the object size change by changing the object depth in weak perspective image, i.e., $\Delta z = (1/s-1)z$ where $z$ and $\Delta z$ are the depth of the object and $\Delta z$ is the change of the depth. $\mathbf{R}\in SO(3)$ is 3D rotation matrix that aligns the object center to the principle coordinate where its roll axis can be chosen arbitrary. 

This object centering eliminates the translation from $SE(3)$, resulting in three degrees of freedom for rotation, i.e., $\mathbf{p}\in SO(3)$. We leverage this object centering to reduce the dimension of the transformation, which allows sampling transformations tractable.

\subsection{Relative Pose Inference} \label{Sec:inference}
In the inference phase, we estimate the relative transformation given the source and target image by minimizing the following objective:
\begin{align}
    \mathbf{p}^* = \underset{\mathbf{p}}{\operatorname{argmin}}~\|f(\mathbf{I}_{\rm tar}) - h_{\mathcal{W}}(f(\mathbf{I}_{\rm src}))\|^2, \label{Eq:inference}
\end{align}
where $\mathbf{p}^*$ is the optimal relative transformation. In practice, we use the gradient decent method to optimize Equation~(\ref{Eq:inference}):
\begin{align}
    \mathbf{p} = \mathbf{p} - \gamma \frac{\partial e}{\partial \mathbf{p}},
\end{align}
where $e = \|f(\mathbf{I}_{\rm tar}) - h_{\mathcal{W}}(f(\mathbf{I}_{\rm src}))\|^2$, and $\gamma$ is the step size.

\subsection{Implementation and Closed-loop Servoing}

We implement the equivariant representation learning using a Siamese network that integrates the feature extractor $f$ and the feature transformer $h$ as shown in Fig. \ref{fig:total_network}. The feature extractor is made of a ResNet18 encoder pre-trained on ImageNet~\cite{he2016deep}, followed by three layers of MLPs with batch normalizations and ReLU activations. The input image size is $224 \times 224 \times 3$, and the output image feature from the input image is a 128-dimensional vector. The weights of two feature extractors (source and target images) are shared. The feature transformer is composed of MLP with two layers with ReLU activation in the first layer. The feature transformer outputs a residual feature for faster convergence. While constraining the transformation with the reduced dimension in Section~\ref{Sec:obj_center}, in practice, we use the over-parametrization of the transformation for the feature transformer, i.e., the transformer takes as input a 7-dimensional vector (three for translation and four for quaternion). For the gradient decent method in Section~\ref{Sec:inference}, we project $\mathbf{p}$ onto the reduced dimension for updates. 


We iterate our feature transformer to further refine the relative transformation by updating the current source image. Specifically, the image captured by the camera on a manipulator is used as a source image, and the inferred transformation is applied to the manipulator by solving inverse kinematics. This process is repeated until it reaches the desired target image. 

\begin{figure}[t]
\centering
\includegraphics[width=\columnwidth]{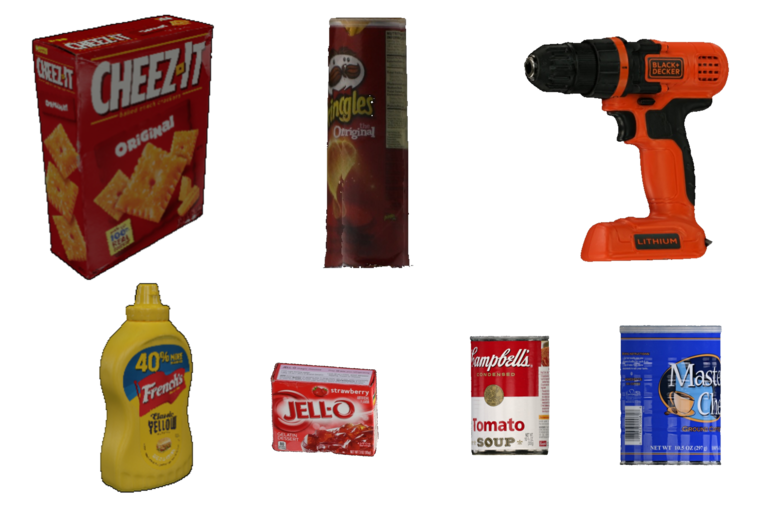}
\caption{YCB objects used in this paper.}
\label{fig:ycb_obj}
\end{figure}

\label{sec:methods}
%

\section{Results}

\begin{table*}[]
\begin{tabular}{l|cl|ll|ll|cl}
\toprule
\multirow{2}{*}{Object}
&
  \multicolumn{2}{c|}{\begin{tabular}[c]{@{}c@{}}\texttt{IBVS}~\cite{chaumette2006visual}\end{tabular}} &
  \multicolumn{2}{c|}{\begin{tabular}[c]{@{}c@{}}\texttt{PoseCNN}+VS~\cite{xiang2017posecnn}\end{tabular}} &
  \multicolumn{2}{c|}{\begin{tabular}[c]{@{}c@{}}\texttt{RPR}+VS\end{tabular}} &
  \multicolumn{2}{c}{\begin{tabular}[c]{@{}c@{}}\texttt{Ours}+VS\end{tabular}} \\
  \cmidrule(lr){2-9}
&
  \multicolumn{1}{c}{PCS$_{0.03}\uparrow$} &
  \multicolumn{1}{c|}{ADD$\downarrow$} &
  \multicolumn{1}{c}{PCS$_{0.03}\uparrow$} &
  \multicolumn{1}{c|}{ADD$\downarrow$} &
  \multicolumn{1}{c}{PCS$_{0.03}\uparrow$} &
  \multicolumn{1}{c|}{ADD$\downarrow$} &
  \multicolumn{1}{c}{PCS$_{0.03}\uparrow$} &
  \multicolumn{1}{c}{ADD$\downarrow$} 
   \\ 
  \hline
Cracker box &
  \multicolumn{1}{c}{0.00} &
  \multicolumn{1}{c|}{0.1246} & 
  \multicolumn{1}{c}{0.64} &
  \multicolumn{1}{c|}{0.0351} &
  \multicolumn{1}{c}{0.46} &
  \multicolumn{1}{c|}{0.0392} &
  \multicolumn{1}{c}{0.91} &
  \multicolumn{1}{c}{\textbf{0.0181}}
   \\ 
Chip can &
  \multicolumn{1}{c}{0.01} &
  \multicolumn{1}{c|}{0.1635} &
    \multicolumn{1}{c}{-} &
  \multicolumn{1}{c|}{-} &
  \multicolumn{1}{c}{0.33} &
  \multicolumn{1}{c|}{0.0457} &
  \multicolumn{1}{c}{0.76} &
  \multicolumn{1}{c}{\textbf{0.0231}}
   \\ 
Gelatin box &
    \multicolumn{1}{c}{0.07} &
  \multicolumn{1}{c|}{0.0568} &
    \multicolumn{1}{c}{0.94} &
  \multicolumn{1}{c|}{0.0236} &
  \multicolumn{1}{c}{0.89} &
  \multicolumn{1}{c|}{0.0172} &
  \multicolumn{1}{c}{0.99} &
  \multicolumn{1}{c}{\textbf{0.0077}}
   \\ 
Master can &
    \multicolumn{1}{c}{0.02} &
  \multicolumn{1}{c|}{0.0832} &
    \multicolumn{1}{c}{0.39} &
  \multicolumn{1}{c|}{0.0417} &
  \multicolumn{1}{c}{0.76} &
  \multicolumn{1}{c|}{0.0282} &
  \multicolumn{1}{c}{1.00} &
  \multicolumn{1}{c}{\textbf{0.0103}}
   \\ 
Mustard bottle &
    \multicolumn{1}{c}{0.02} &
  \multicolumn{1}{c|}{0.0749} &
    \multicolumn{1}{c}{0.13} &
  \multicolumn{1}{c|}{0.0457} &
  \multicolumn{1}{c}{0.79} &
  \multicolumn{1}{c|}{0.0227} &
  \multicolumn{1}{c}{1.00} &
  \multicolumn{1}{c}{\textbf{0.0074}}
   \\ 
Power drill &
    \multicolumn{1}{c}{0.00} &
  \multicolumn{1}{c|}{0.1389} &
    \multicolumn{1}{c}{0.35} &
  \multicolumn{1}{c|}{0.0396} &
  \multicolumn{1}{c}{0.44} &
  \multicolumn{1}{c|}{0.0423} &
  \multicolumn{1}{c}{0.91} &
  \multicolumn{1}{c}{\textbf{0.0160}}
   \\ 
Tomato can &
    \multicolumn{1}{c}{0.02} &
  \multicolumn{1}{c|}{0.0572} &
    \multicolumn{1}{c}{0.76} &
  \multicolumn{1}{c|}{0.0313} &
  \multicolumn{1}{c}{0.92} &
  \multicolumn{1}{c|}{0.0159} &
  \multicolumn{1}{c}{1.00} &
  \multicolumn{1}{c}{\textbf{0.0070}}
   \\ \bottomrule
\end{tabular}
\caption{We compare our method with state-of-the-art algorithms using ADD and PCS$_{0.03}$ metrics, i.e., the ADD tolerance is 3cm, which is acceptable tolerance for an object manipulation task. Ours produces the highest PCS and the lowest ADD.} 
\label{table:comparison_result}
\end{table*}

\begin{figure*}
\begin{multicols}{3}
    \includegraphics[width=\linewidth]{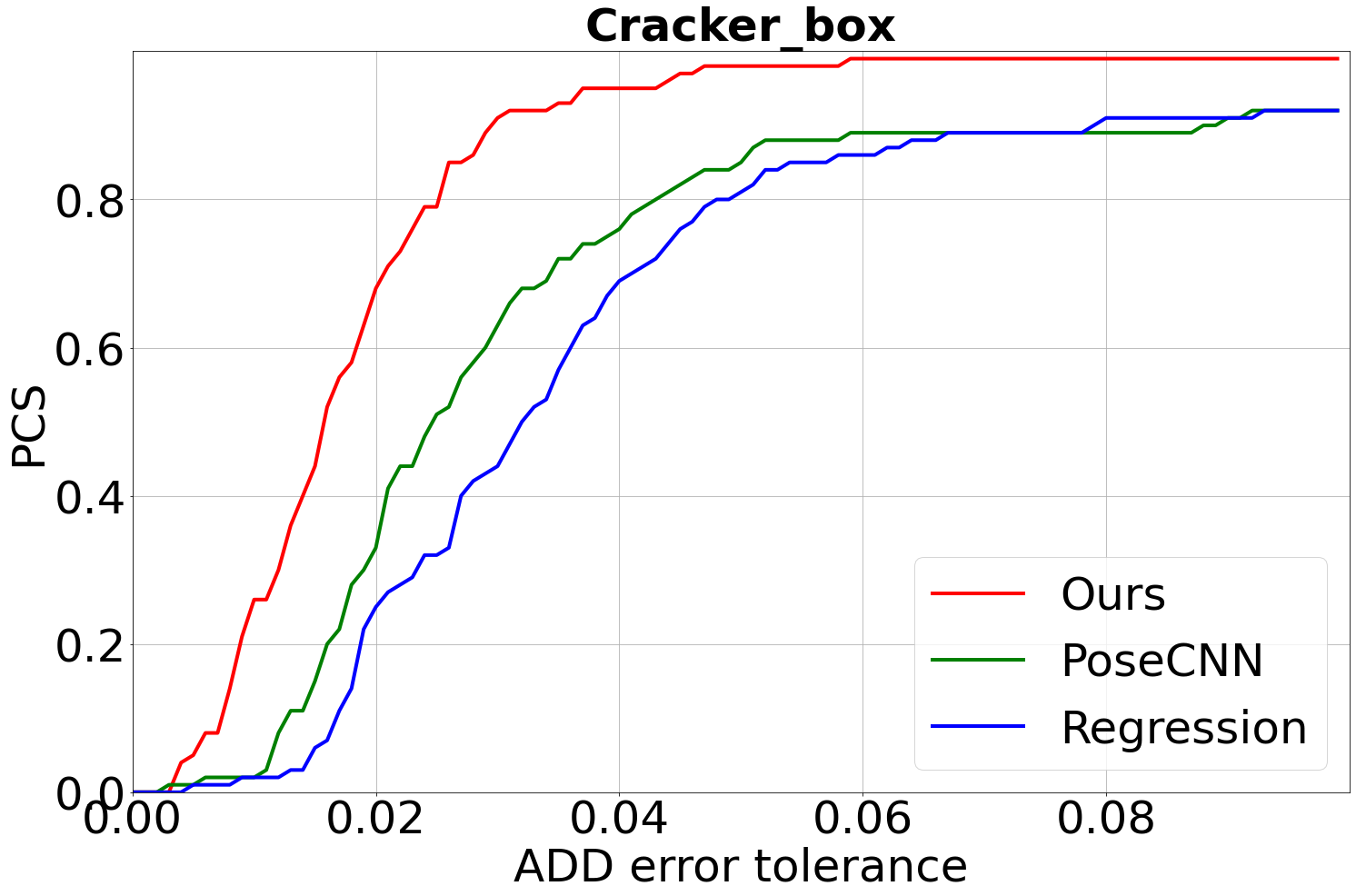}\par 
    \includegraphics[width=\linewidth]{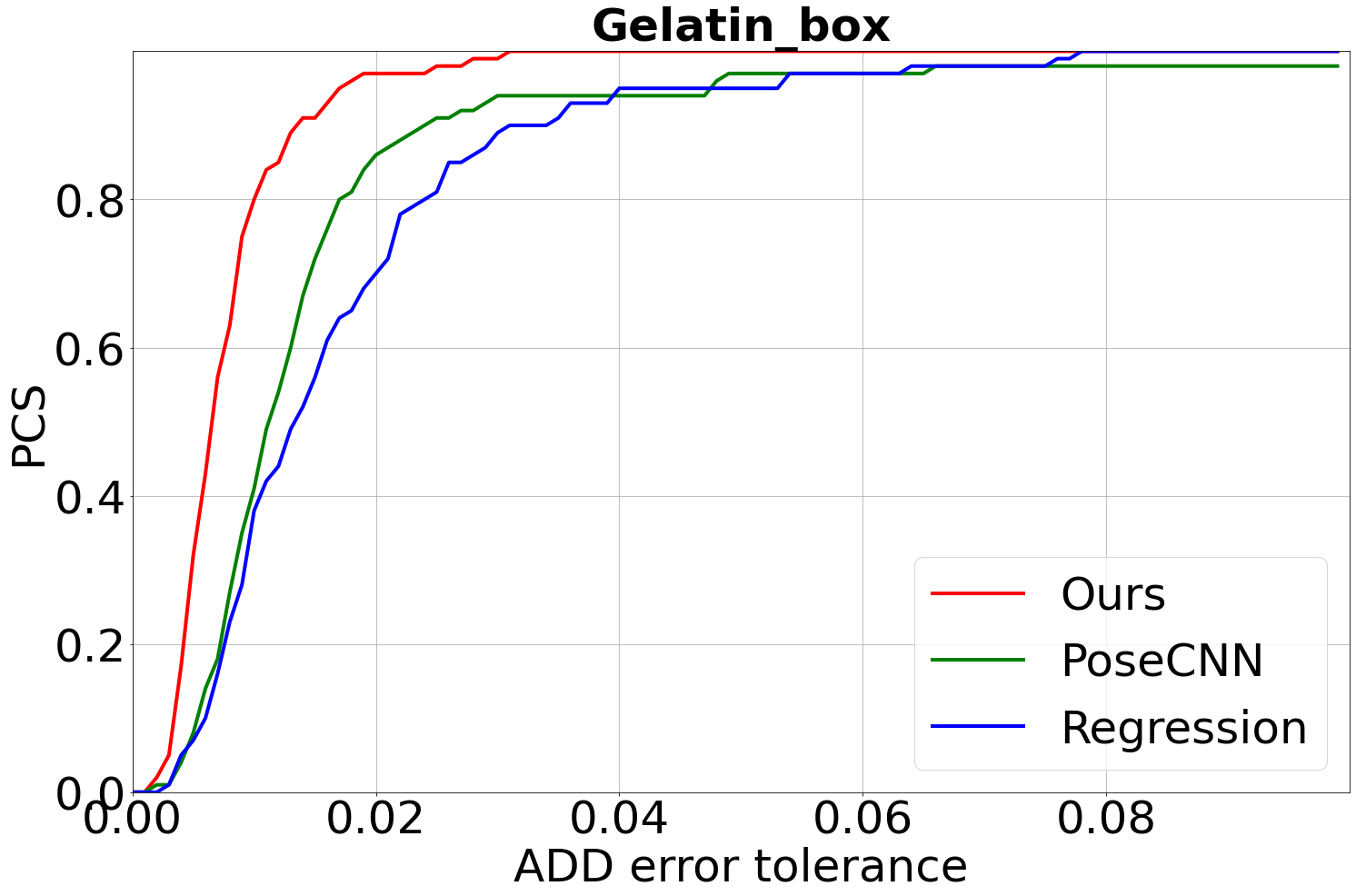}\par 
    \includegraphics[width=\linewidth]{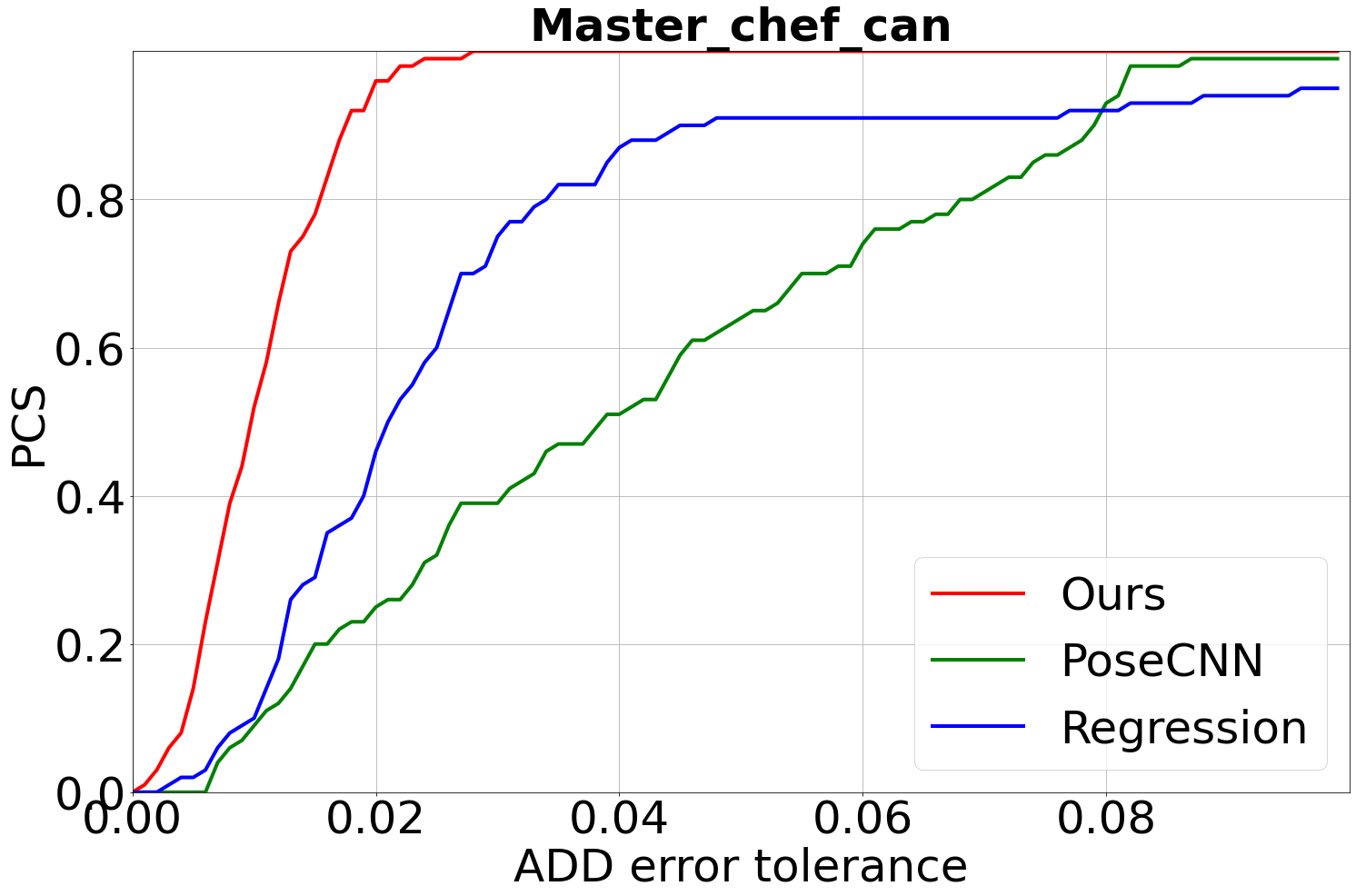}\par 
    \end{multicols}
\begin{multicols}{3}
    \includegraphics[width=\linewidth]{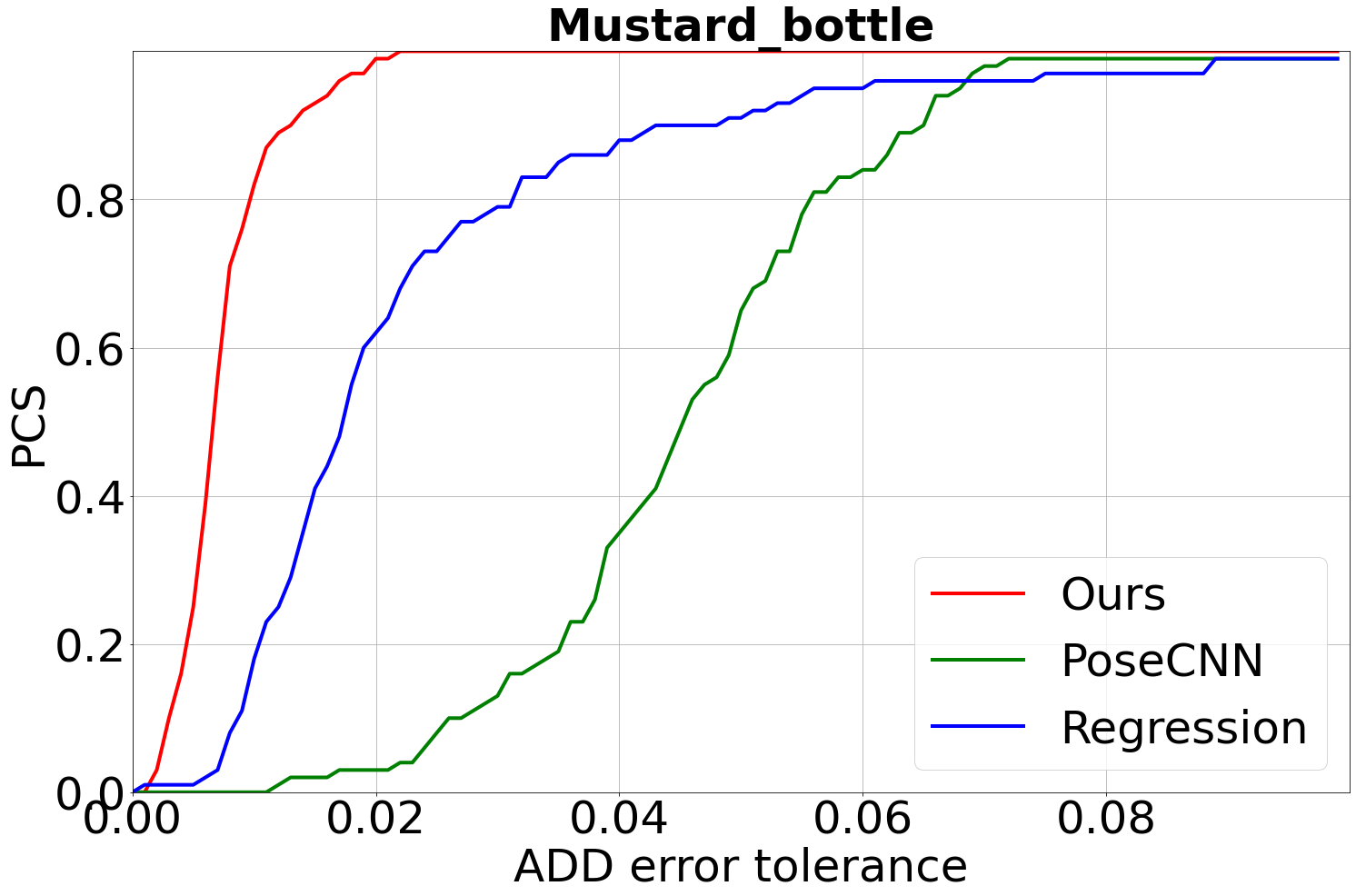}\par
    \includegraphics[width=\linewidth]{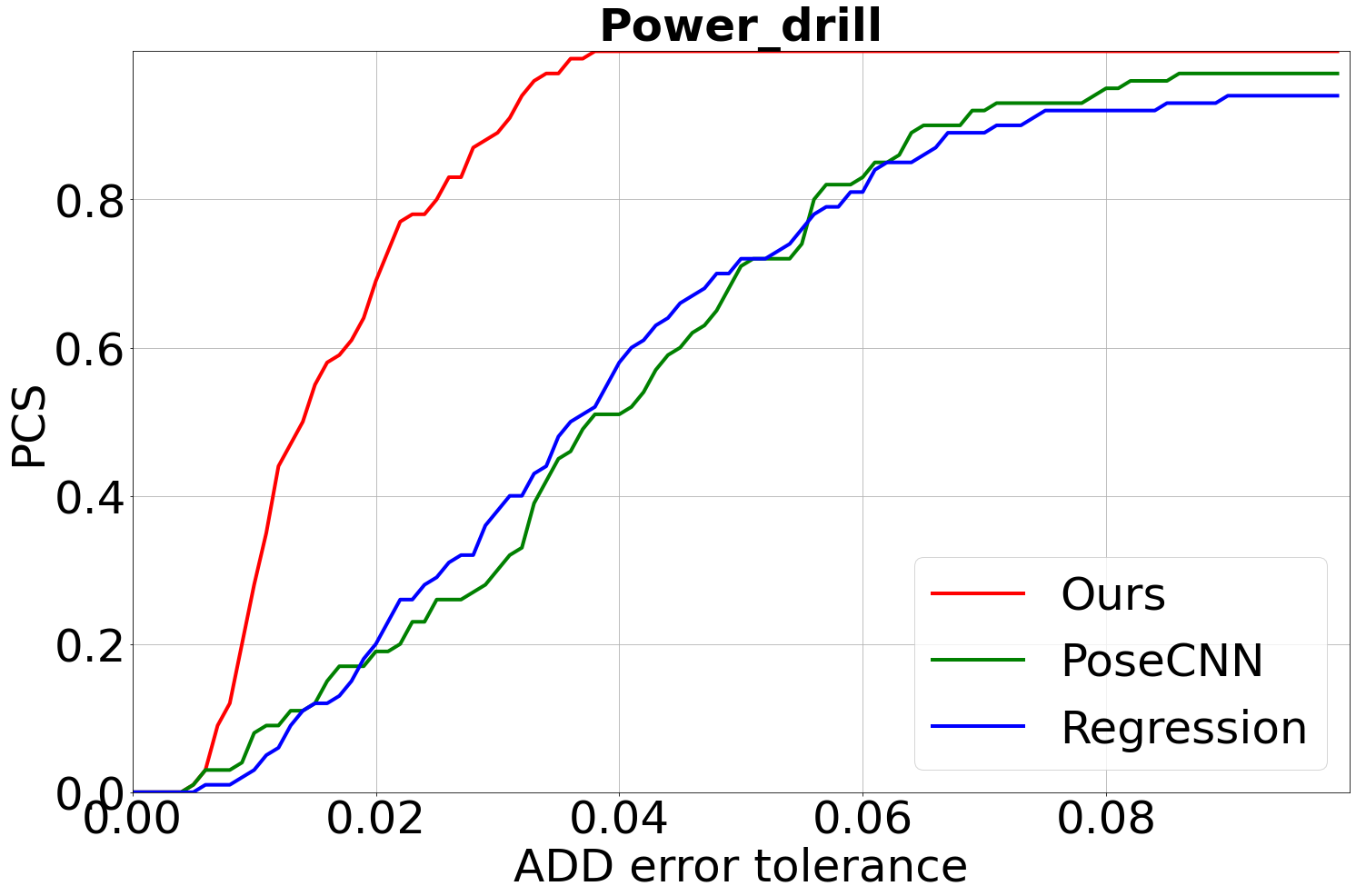}\par
    \includegraphics[width=\linewidth]{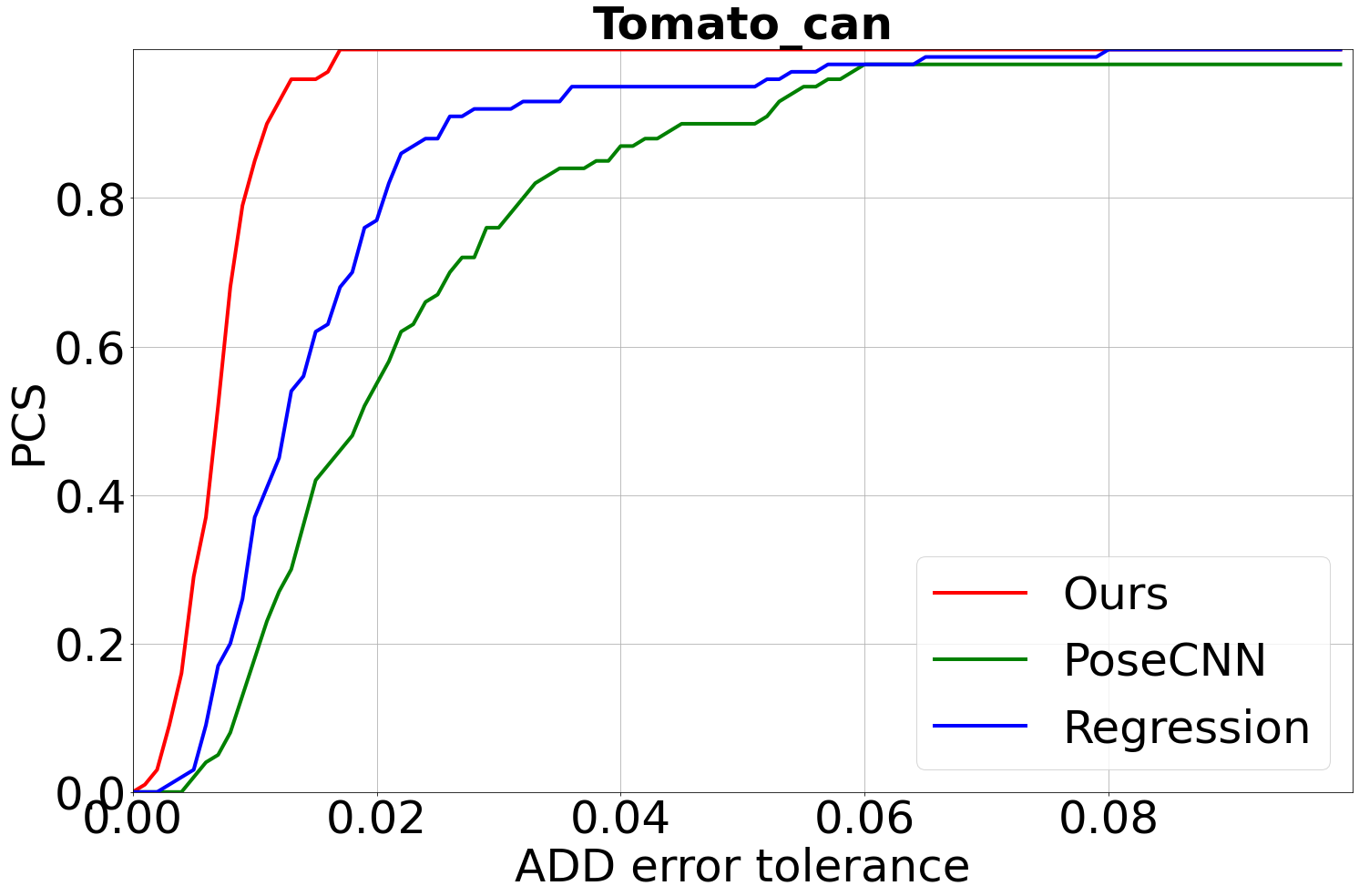}\par 
\end{multicols}
\caption{We show full PCS as varying ADD tolerance. Ours significantly outperforms the others with a large margin.}
\label{fig:result_add}
\vspace{1mm}
\end{figure*}

In this section, we evaluate our method on the visual servoing task by comparing it with state-of-the-art learning-based approaches. For training, we use $140,000$ images ($20,000$ images per object) rendered from random viewpoints using YCB $3$D models with partial occlusions as shown in Fig. \ref{fig:ycb_obj}. Given the generated images, we make $2,800,000$ image pairs with their transformations. At each training epoch, we randomly sample $140,000$ images pairs and corresponding transformations.

\subsection{Baseline Algorithms and Metrics}
We compare our method with three baseline approaches. 1) \texttt{IBVS}: Image-based visual servoing~\cite{chaumette2006visual} takes the source and target images as input and outputs relative poses given camera intrinsic parameters. It matches local image features such as SIFT~\cite{lowe2004distinctive} using a parametric transformation, e.g., homography. Given the matches, it estimates the $SE(3)$ transformation approximated by the pseudo inverse of the interaction matrix $\mathbf{L}_{e}^{+}$ in Equation~(\ref{eq:cam_vel})~\cite{chaumette2006visual} with the constant depth value. 
\begin{equation}\label{eq:cam_vel}
    \mathbf{v}_{c} =
    \begin{bmatrix}
    v_{c} \\
    \omega_{c}
    \end{bmatrix}
    = -\lambda_e \mathbf{L}_{e}^{+}\mathbf{e}
\end{equation}
where $\lambda_e$ is a hyperparmeter that we decide empirically from our experiments, and $\mathbf{e}$ is the error in image plane between the filtered matching features. $v_{c}$ and $\omega_{c}$ are linear and angular velocity of the camera in 3D. 2) \texttt{PoseCNN}: PoseCNN~\cite{xiang2017posecnn} is a supervised learning method that regresses the absolute camera pose with respect to the pre-defined object coordinate system. It requires 3D bounding box ground truth of an object to train. It is trained on YCB 3D models. We use the pre-trained model for the target pose estimation, i.e., no relative transformation is estimated. 3) \texttt{RPR}: We design a baseline that directly regresses the relative transformation without 3D ground truth (relative pose regression). A key difference from ours is that it lacks the feature transformer where 3D equivariance is not applied. We use the architecture design proposed by \cite{bateux2017visual} with ResNet18 and train it on our dataset. 




 

We use two evaluation metrics to measure performance. 
(1) ADD: Average distance metric~
\cite{xiang2017posecnn,hinterstoisser2012model,tremblay2018deep} is 3D Euclidean distance of pairwise points between the 3D model point clouds transformed by the ground truth and the estimated pose as follows: 
\begin{align}
\text{ADD} = \frac{1}{|\mathcal{M}|} \sum_{\textbf{x} \in \mathcal{M}} ||(\textbf{R}\textbf{x}+\textbf{T}) -(\tilde{\textbf{R}}\textbf{x}+\tilde{\textbf{T}})||
\end{align}
where $\mathcal{M}$ is a set of point clouds in the model and $|\mathcal{M}|$ is the cardinality of the set. \textbf{R} and \textbf{T} are the ground truth absolute pose of the image with respect to the object coordinate frame, and $\tilde{\textbf{R}}$ and $\tilde{\textbf{T}}$ are the estimated pose. For the evaluation and comparison with others, although our method estimates a relative transformation, we compute the estimated pose of the target image with the estimated transformation. The lower, the better. (2) PCS: We design a new metric called Probability of Correct Servoing that measures successful servoing rate:
\begin{align}
    \text{PCS}_{\epsilon} = \frac{\sum_{i}\delta(\text{ADD}_i < \epsilon)}{N},
\end{align}
where $\delta$ is the Dirac delta function, and $\text{ADD}_i$ is the ADD of the $i^{\rm th}$ testing data instance. $\epsilon$ is the error tolerance to operate successful servoing for object manipulation tasks, and $N$ is the number of testing data instances. The higher, the better.



\subsection{Comparison Results}
Table \ref{table:comparison_result} shows the comparison results on ADD and PCS metrics. 
\texttt{RPR} is less accurate in PCS with higher ADD than ours. This is mainly due to the lack of 3D equivariance in the learned representation, which is likely to be overfitted to the training data. 
In terms of ADD, ours achieves $36$\% of ADD compared to \texttt{PoseCNN} and $43$\% of ADD compared to \texttt{RPR}. Fig.~\ref{fig:result_add} illustrates the full PCS measure over the error tolerance where ours achieves higher PCS with a lower error tolerance.

\begin{figure}
    \centering
    \includegraphics[width=\textwidth]{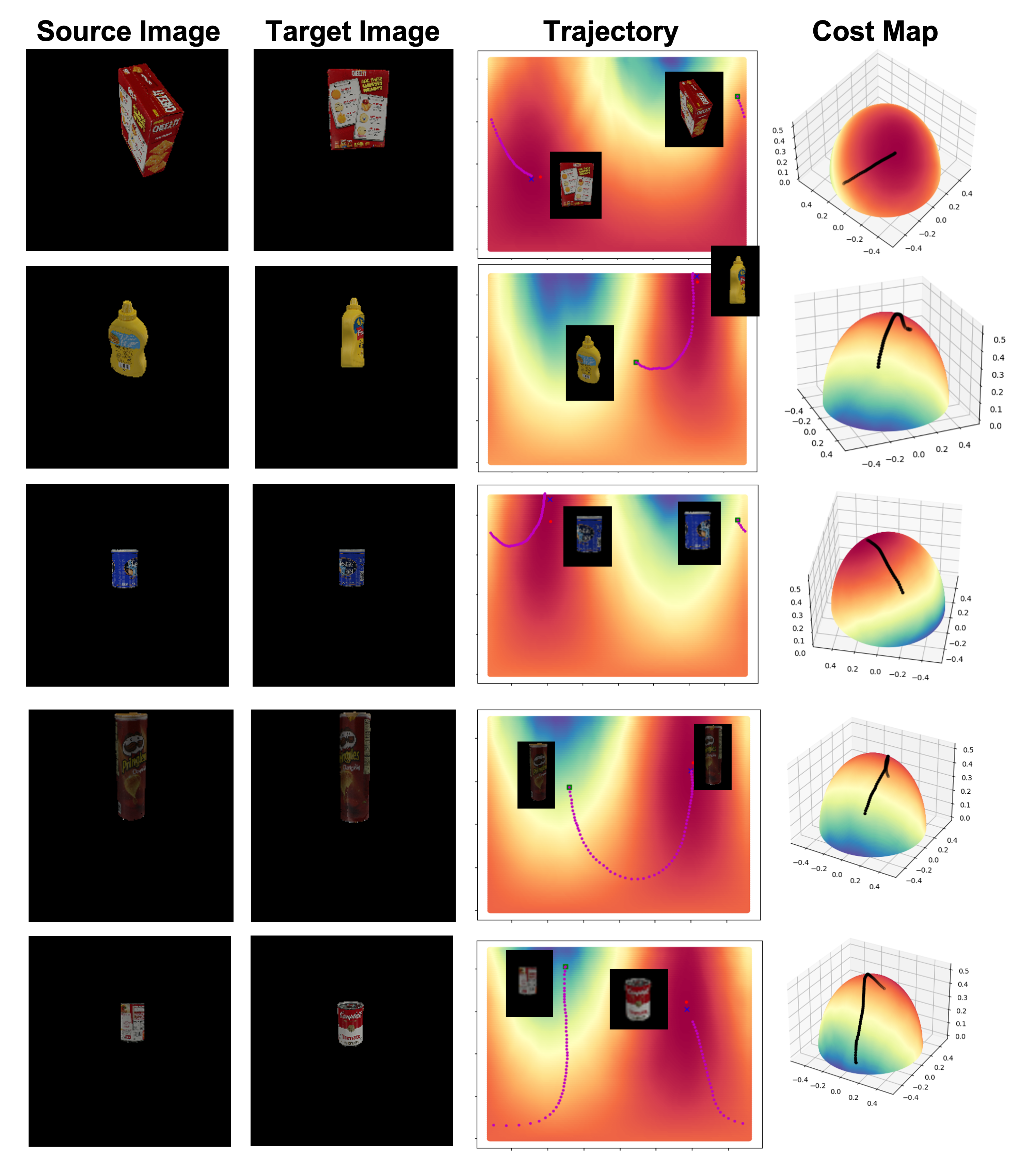}
    \caption{We illustrate visual servoing over the wide baseline source and target images. The trajectory in the cost map shows feature distance between the source and current image, which is reflective of 3D geodesic distance, i.e., reduced $SE(3)$. }
    \label{traj_images}
\end{figure}



\subsection{Ablation Study for Object Centering}
\label{section:ablation_study}
\begin{table}[t]
\small
\begin{tabular}{cccc}
\toprule
Object &
  \begin{tabular}[c]{@{}c@{}}without \\ object \\ centering\end{tabular} &
  \begin{tabular}[c]{@{}c@{}} with \\ object \\ centering\end{tabular} &
  \begin{tabular}[c]{@{}c@{}}Reduction\end{tabular} \\ \hline
Cracker box    & 0.1920 & 0.0703 & -63.38\% \\ 
Chip can       & 0.2372 & 0.1348 & -43.17\% \\ 
Gelatin box    & 0.3220 & 0.0717 & -77.73\% \\ 
Master can     & 0.2721 & 0.0950 & -65.09\% \\ 
Mustard bottle & 0.2020 & 0.0734 & -63.66\% \\ 
Power drill    & 0.2069 & 0.1027 & -50.36\% \\ 
Tomato can     & 0.2799 & 0.1184 & -57.69\% \\ \bottomrule
\end{tabular}
\caption{We leverage the object centering scheme to reduce the dimensionality of $SE(3)$, which allows learning tractable. Our method with centering yields more than 50\% of ADD reduction. }
\label{table:ablation_result}
\end{table}

We use the object centering scheme for dimensionality reduction, which allows us to learn a coherent representation. We evaluate the effectiveness of object centering. 
For the one w/o object centering, we learn all possible transformation in $SE(3)$ directly from sampled source and target images. On the other hand, the one w/ object centering pre-applies the homography in Equation~(\ref{Eq:homography}) to the source image as if the camera is closely tracking the object with a fixed distance. We learn the representation from the object-centered images. After applying the predicted relative transformation, we post-apply object de-centering to form the target image. Table~\ref{table:ablation_result} summarizes the performance in ADD metric. 
Note that the method with object centering is trained with the dataset from hemisphere space, and inputs images generated by homography. Although it is trained with a different dataset from the testing dataset, it shows more robust performance than the one without object centering which learns the network directly in $SE(3)$.



\subsection{Visual Servoing on Real Robotic Arm}
\begin{figure}[t]
\centering
\begin{subfigure}[b]{0.47\columnwidth}
\includegraphics[width=\textwidth]{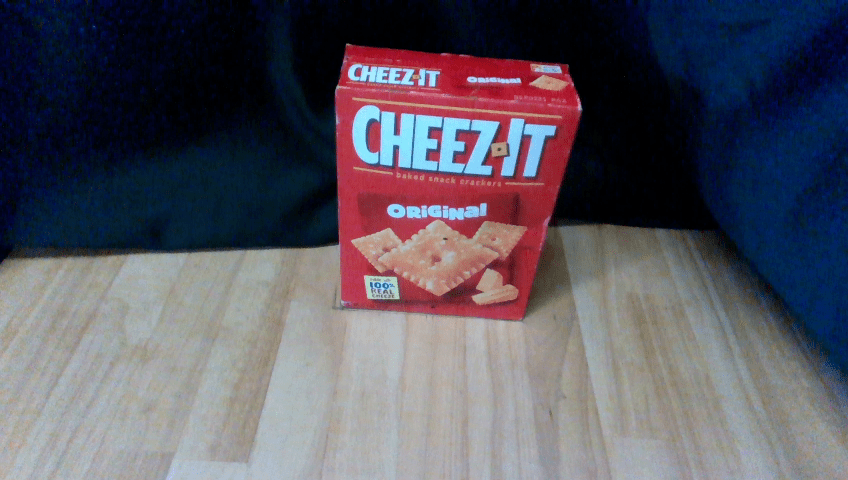}
\caption{Target image} \label{fig:target_real_img}
\end{subfigure}
\begin{subfigure}[b]{0.47\columnwidth}
\includegraphics[width=\textwidth]{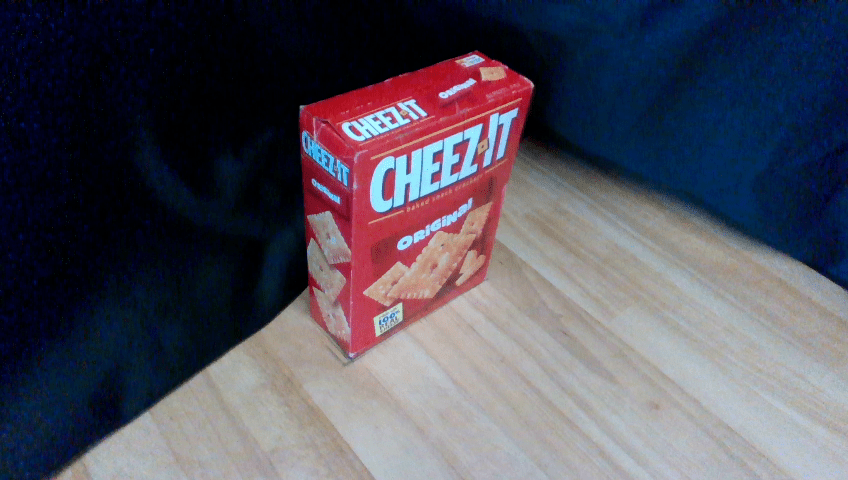}
\caption{Current image} \label{fig:current_real_img}
\end{subfigure}
\begin{subfigure}[b]{0.47\columnwidth}
\includegraphics[width=\textwidth]{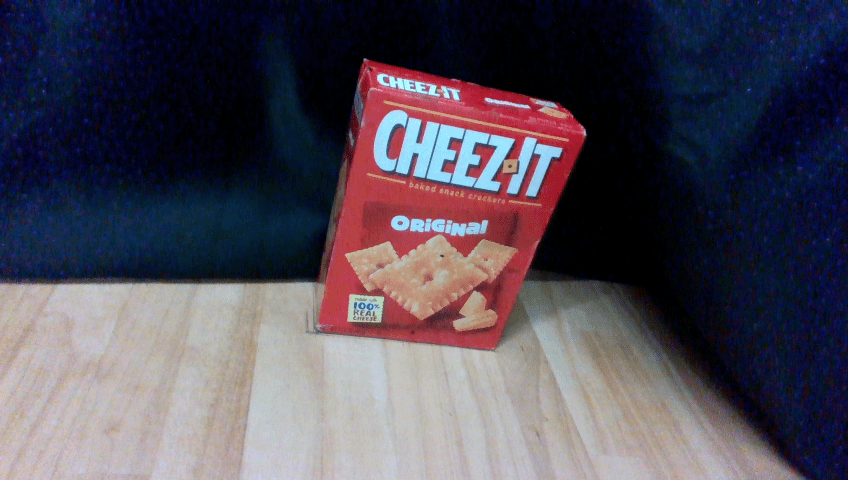}
\caption{Image using network} 
\label{fig:network_real_img}
\end{subfigure}
\begin{subfigure}[b]{0.47\columnwidth}
\includegraphics[width=\textwidth]{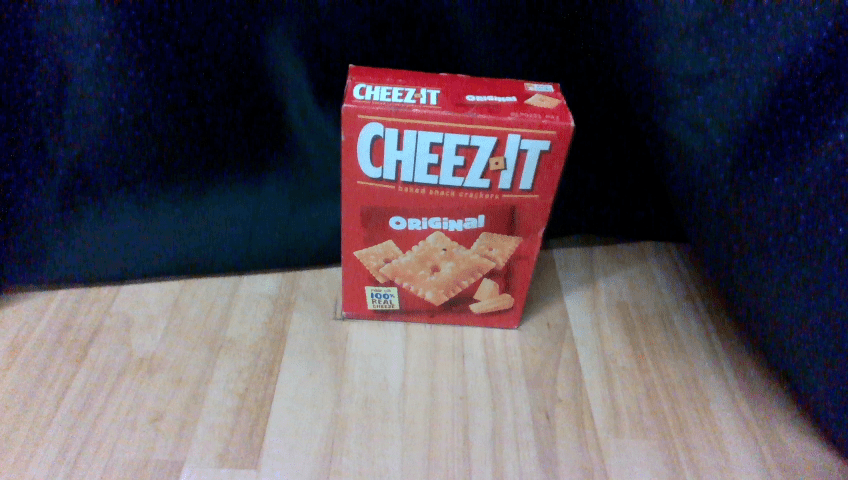}
\caption{Final image} \label{fig:final_real_img}
\end{subfigure}
\caption{Ours generates a trajectory of relative transformation from the current image to the target image. (c) shows the image based on the estimated pose by our network method and (d) shows the final image by visual servoing.}
\label{fig:real_exp}
\vspace{-3mm}
\end{figure}

To verify the performance of our method in a real environment, we conducted an experiment with a robot manipulator and an Intel RealSense camera on the wrist of the robot. Our method generates a trajectory of relative transformations to reach a given target image (Fig.~\ref{fig:target_real_img}) from a current image (Fig.~\ref{fig:current_real_img}). We used MaskRCNN~\cite{he2017mask} for segmented images for input of the network. Fig.~\ref{fig:network_real_img} shows the image after following the trajectory of relative transformations generated by the network, and Fig.~\ref{fig:final_real_img} shows the final image by visual servoing. Fig.~\ref{traj_images} shows errors of camera poses from the initial camera pose to the final camera pose following the network gradient and visual servoing. The final camera pose errors are less than $5$cm and $5$ degrees.

\begin{figure}
    \centering
    \includegraphics[width=\textwidth]{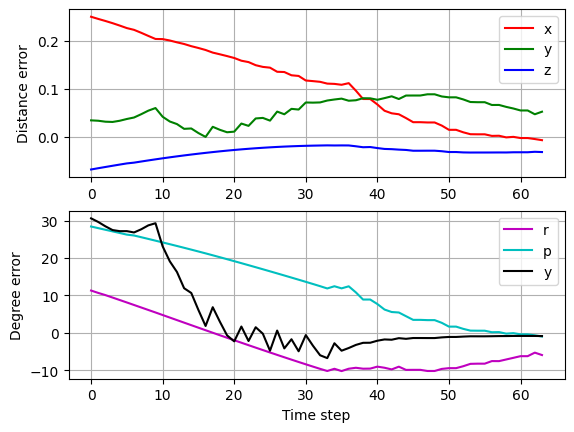}
    \caption{The errors of the camera pose with respect to the target camera pose. }
    \label{traj_images}
\end{figure}



\section{Conclusion}
This paper presents a new self-supervised learning method to learn a visual representation for a wide baseline servoing task without 3D supervision. We leverage 3D equivariance that enforces a geometric constraint on the visual representation, i.e., the representation must be transformed as a function of 3D camera egomotion transformation. With 3D equivariance, the visual representation and its transformation are jointly learned via a Siamese network made of a feature extractor and feature transformer. To avoid a trivial solution, we incorporate a geodesic preserving constraint. With the learned representation, we formulate an inference method to estimate the optimal transformation that can best explain the visual feature transformation. Further, to address the challenge of sampling over high dimensional space, we introduce an object centering scheme that can effectively reduce the space of $SE(3)$. The results show strong performance compared to state-of-the-art visual servoing methods, quantitatively. Our method takes segmented images as input therefore it is agnostic to the scene complexity. However, the current version was evaluated under no or mild occlusion conditions. In the future, we would like to extend our method to complex scenes where there might be significant occlusions caused by other objects in the scene.

\addtolength{\textheight}{-3.7cm}  

\bibliographystyle{IEEEtran}
\bibliography{main.bib}


\end{document}